# Acting on the Unseen: Communication-Free Collaborative Filtering for Decentralized Multi-Robot Task Allocation

Alexander Apartsin[a], Yigal Meshulam[b], Yehudit Aperstein[b]

[a]Holon Institute of Technology, Holon, Israel.

[b]Afeka Tel Aviv Academic College of Engineering, Tel Aviv, Israel

**Abstract.** Multi-robot task allocation usually assumes some combination of communication, known task models, or a coordinator. We study the opposite extreme, a regime common in practice but overlooked in theory, which we name **Zero-Knowledge MRTA** (ZK-MRTA): a robot team with **no prior knowledge** (no task models, not even the latent rank), **no communication** (no messages, no parameter sharing, no coordinator), and only a **partial and privately-noisy** view of a public stream of teammates' outcomes. A hidden low-rank structure governs which robot suits which task, and there are far more tasks than rounds, so most (robot, task) pairs are never attempted. Yet each robot can act well on tasks it never attempted, and onboard new tasks, by running online low-rank collaborative filtering over the broadcast (**SwarmCF**). The advantage over any structure-free learner is **categorical**, not a constant factor: a structure-free learner is provably at the prior-mean error floor on unseen pairs. We prove a matching per-robot sample complexity ($\Theta(d)$ versus $\Theta(n)$, in the rank $d$ and the task count $n$), an anytime (cumulative-reward) separation under task scarcity, and a deterministic condition under which decentralized recovery from the masked broadcast is exact (validated empirically). Experiments quantify the value of the broadcast, a positive scaling law (per-robot unseen-pair skill rises with team size), and the strongest masking-robustness and anytime profile among low-rank methods, recovering most (about 80% on earned skill) of a centralized full-communication ceiling, and holding under capacity-1 contention and in a robotics-grounded sensing instance.

## 1. Introduction

Consider a team of autonomous robots, for example aerial vehicles, that must repeatedly decide which task to engage: which area to inspect, which target to service, which sensor reading to pursue. Whether a given robot does well on a given task depends on a match between the robot's **capabilities** (sensor modalities, payload, effector type, endurance, remaining consumables) and the task's **requirements**. This capability-requirement view is the basis of trait-based task allocation [1], and the match is typically governed by only a few underlying factors, so the full robot × task reward is **low-rank**. That a few factors suffice is borne out in practice: learned low-dimensional capability and task-requirement models predict real robotic task performance to within a few percent [2], and the analogous two-sided affinity matrix is empirically low-rank in large matching datasets [3].

Most multi-robot task allocation (MRTA) methods obtain coordination from at least one of three resources: explicit **communication** (auctions, consensus, message-passing), **known task models or utilities**, or a **central coordinator / centralized training**. In many real deployments none of these is available: communication is jammed, bandwidth-limited, or deliberately withheld for stealth; the task structure is unknown a priori; and there is no coordinator. What a robot can often still do is **passively sense** some of what its teammates are doing and how it turned out, imperfectly, at a distance, and differently from every other robot. This is documented, not hypothetical: fielded drone swarms already coordinate from purely onboard, range- and line-of-sight-limited mutual perception with no inter-robot messaging [4,5], and heterogeneous teams routinely operate under denied or degraded communication, as across the DARPA Subterranean Challenge [6,7]. This is the regime we formalize and solve: a robot observes a **partial** (range-limited) and **privately-noisy** slice of a public outcome stream, never the same slice as a teammate.

The technical crux is **generalization to the unseen**. There are far more tasks than rounds ($n \gg T$), so each robot personally attempts only a vanishing fraction of tasks. A learner that estimates each task only from its own attempts (a **structure-free** learner: independent per-task bandits, tabular value tables) has, on any task it never attempted, nothing better than the prior mean, the error floor, which dominates exactly because tasks outnumber rounds. The opportunity is that the shared low-rank structure links tasks: outcomes a robot *observes* teammates obtain (even noisily and partially) constrain that structure, and a few observations then determine the robot's reward on tasks it has never touched. We show a single, simple estimator turns this opportunity into a categorical capability.

**The gap.** To our knowledge no prior method targets this cell. Every established paradigm relaxes at least one of its defining constraints (no prior knowledge, no communication, decentralized decisions, and partial and privately-noisy observation) by assuming communication, known utilities/traits, a coordinator or centralized training, or clean/shared observation (Section 2, Table 1). The regime is not exotic: it is the default when communication is jammed, bandwidth-limited, or withheld for stealth, and when task structure must be discovered in the field. We close it.

**Relation to collaborative filtering.** Low-rank collaborative filtering is itself classical; our contribution is not the estimator but the demonstration, with theory, that it works *at all* in this regime, fully decentralized, communication-free, and under a persistent, per-observer-private observation mask where standard uniform-sampling completion guarantees do not apply, together with the recovery condition that says exactly when it works and the collective-speedup law that says why a swarm helps. The categorical floor against structure-free learning, the decentralized recovery condition under a structured private mask, and the value-of-broadcast / positive-scaling results are, to our knowledge, new.

**Contributions.**

1. We formalize and name **Zero-Knowledge MRTA** (ZK-MRTA): communication-free MRTA under partial, privately-noisy observation and zero prior knowledge, a most-restrictive but practically common regime that prior MRTA and decentralized-learning work does not address (Section 2, Table 1).
2. We propose **SwarmCF**, a decentralized, online, low-rank collaborative filter that each robot runs over the passive broadcast, with a constant-time **fold-in** (an $O(\hat{d}^3)$ ridge solve in the guessed rank $\hat{d}$, independent of the task count $n$; Section 4) that lets it act on unseen tasks and onboard new tasks.
3. We prove the advantage is **categorical**: a structure-free learner is at the error floor on unseen pairs and the broadcast is provably uninformative to it, whereas SwarmCF attains a per-robot sample complexity of $\Theta(d)$ versus $\Theta(n)$, with a matching **anytime** separation under task scarcity (Section 5).
4. We give a **deterministic condition** under which decentralized recovery of the shared structure from the privately-masked broadcast is exact, with a coverage-time bound that improves as the team grows, and validate the condition empirically (Section 5, Appendix C).
5. We quantify the **value of the broadcast**, a **positive scaling law** (per-robot competence rises with team size), and the **strongest masking-robustness and anytime profile** among low-rank methods under limited observability, recovering most of a centralized full-communication ceiling on earned (anytime) skill; and we show the advantage **persists under capacity-1 contention** (Section 6.5) and in a **robotics-grounded instance** with heterogeneous sensing traits and a range-limited line-of-sight mask (Section 6.7).

6. We release **LatentSwarm**, an open, modular Python package for ZK-MRTA, covering the masked-broadcast setting used for our headline results and its capacity-1 contention extension; with it we show the separation survives contention and that SwarmCF's heterogeneous per-robot models **implicitly de-conflict** where structure-free learners instead collide, isolating communication-free de-confliction as the main open problem (Section 6.5, Appendix E).

## 2. Related work

Swarm robotics seeks collective competence from simple local rules [8,9], and coverage and patrolling control coordinate where robots *move* [10]; we instead address which *task* each robot should engage when the capability-to-task match is unknown and there is no communication.

**Multi-robot task allocation.** The taxonomies of Gerkey and Matarić [11], Korsah et al. [12], and Nunes et al. [13] organize MRTA by single/multi-task robots, single/multi-robot tasks, and instantaneous/time-extended assignment with interrelated utilities and temporal constraints. Classical solvers, market and auction mechanisms [14], consensus-based bundle algorithms (CBBA) [15], and distributed constraint optimization achieve coordination through **communication** and assume **known** task utilities or costs (surveys: [16]); even work that explicitly limits communication still relies on auctions and messages [17], and a recent review underscores how central communication remains to multi-robot systems [18]. Trait-based and heterogeneity-aware MRTA [1,19,20] matches robot capability vectors to task requirement vectors, but takes the traits as **given**. Our setting keeps the trait/low-rank view but makes the traits **unknown and learned online**, with neither communication nor known utilities; recent surveys [21] document the rapid growth of learning-based MRTA, but the prior-free, communication-free regime we study remains unaddressed.

**Decentralized and learning-based coordination.** Communication-free multiplayer bandits (musical chairs [22], SIC-MMAB [23], game-of-thrones [24]) break symmetry without messages but are **structure-free** (per-arm), so they cannot generalize to unseen arms. Cooperative multi-agent RL (CTDE: MAPPO [25], QMIX [26], VDN [27], MADDPG [28]) and learned-communication methods (CommNet [29], TarMAC [30], DIAL [31]) rely on centralized training or message passing; recent learning-based decentralized assignment (graph neural networks for goal assignment [32] and scheduling [33]) likewise presumes communication or centralized training. Federated and gossip collaborative filtering [34], and federated learning more broadly [35], coordinate by exchanging model parameters or gradients; we exchange nothing but a passively-sensed outcome stream. Decentralized partially-observable control (Dec-POMDPs [36]) addresses long-horizon coordination under shared latent-state dynamics; our rounds are one-shot offered-set choices with no shared state, so the challenge is cross-task generalization from a privately-masked stream rather than long-horizon credit assignment.

**Low-rank estimation and bandits.** Matrix completion gives centralized recovery guarantees under (near-)uniform sampling [37,38,39,40], with practical factorization estimators (matrix factorization [41], Bayesian PMF [42], soft-impute [43], implicit-feedback ALS [44], ranking CF [45]); low-rank and bilinear bandits (explore-then-spectral [46], bilinear [47], generalized-linear [48], clustering-of-bandits [49]) are centralized and/or phase-structured. Decentralized matrix completion [50] distributes the factorization across nodes but still exchanges factors or residuals over a connected communication graph; we forbid all such exchange and recover from passive observation alone. We make estimation decentralized, online, broadcast-only, and robust to a structured (non-uniform) per-robot observation mask, with the unseen-pair error floor turning the gap into a categorical, rather than constant-factor, separation.

Table 1 places the major paradigms on the four axes that define our problem; each established family relaxes at least one axis we hold fixed, and our cell, low-rank with only a guessed rank, no communication, decentralized, masked and noisy, is the one left open.

***Table 1.*** *Established paradigms versus our regime across prior knowledge, communication, decision locus, and observation; the final column names the constraint each relaxes (ours relaxes none).*

| Paradigm | Prior knowledge | Communication | Decisions | Observation | Constraint it relaxes vs. ours |
|---|---|---|---|---|---|
| Auction / consensus / DCOP MRTA [14,15] | task utilities or costs | message-passing | decentralized | full task info | needs communication and known utilities |
| Cooperative MARL (CTDE) [25,26] | none (learned) | central training or messages | central train, decentral. exec | full (in training) | needs a central critic or messages |
| No-communication multiplayer bandits [22,23] | none (per-arm) | none | decentralized | own pulls + collisions | structure-free: no unseen generalization |
| Low-rank completion / bandits [37,46] | low-rank | centralized | centralized | partial (uniform) | centralized and/or explore-then-commit |
| Federated / gossip CF [34,35] | low-rank | parameter exchange | decentralized | partial | shares parameters, not a passive stream |
| Trait-based MRTA [1,20] | known traits | varies | decentralized | full | traits given, not learned |
| **Ours: ZK-MRTA (this paper)** | low-rank, guessed rank only | none (passive sensing) | decentralized | masked + per-observer noisy | none: the open cell (most constrained) |

## 3. Problem setting: Zero-Knowledge MRTA (ZK-MRTA)

(The term "zero-knowledge" here denotes the absence of any prior task knowledge, no task models and not even the latent rank; it is unrelated to the cryptographic notion of zero-knowledge proofs.)

**Reward.** A team of $m$ robots faces $n$ tasks. Robot $i$ has a hidden capability vector $p_i \in \mathbb{R}^d$ and task $j$ a hidden requirement vector $u_j \in \mathbb{R}^d$; the expected reward of robot $i$ engaging task $j$ is their inner product

$$R_{ij} = \langle p_i, u_j \rangle, \qquad R = PU^\top \in \mathbb{R}^{m\times n}, \ \mathrm{rank}(R) = d \ll \min(m,n).$$

The low rank $d$ encodes that only a few traits govern fit. The team does **not** know $P$, $U$, or even $d$ (it uses a guessed rank $\hat{d}$, drawn at random per run, to which the method is robust; Figure 7). We take the reward in normalized form, scaling the latent traits so that $R_{ij}$ is bounded and zero-mean across tasks; this is the bounded, normalized reward referenced by Proposition 1. Zero-mean makes the no-information prior-mean predictor the correct control against which the categorical floor is measured; the degenerate $d = 1$ case, where a shared popularity/bias baseline already suffices, is treated in Section 6.8.

**Interaction.** Each round $t = 1, \dots, T$ every robot $i$ is offered a menu $S_{it} \subseteq [n]$ of tasks (the model permits the full menu of all $n$ tasks; to control scarcity the body experiments offer a uniform random size-$c$ subset, $c = 20$, with the all-tasks menu $c = n$ studied in Appendix F), selects one $a_{it} \in S_{it}$, engages it, and earns $R_{i,a_{it}}$. The operating regime is **task-scarce**: $n \gg T$, so each robot personally engages only $O(T)$ of the $n$ tasks.

**The observation channel (central to the setting).** There is no communication. Each robot instead **passively senses** a public stream of engagement outcomes, but only partially and noisily, and **privately**: **(persistent partial visibility)** robot $i$ observes teammate $k$'s engagements only if $M_{ik} = 1$, where $M_{ik} \sim$ Bernoulli$(\rho)$ is fixed for the whole mission ($M_{ii} = 1$); **(private per-observer noise)** when $i$ observes the outcome of action $a_{kt}$ it reads $R_{k,a_{kt}} + \eta_{ikt}$ with $\eta_{ikt} \sim \mathcal{N}(0, \sigma^2)$ drawn independently per observer, so the same action is read differently by different robots. A robot reads its own engaged outcome with a smaller own-observation noise $\sigma_{\text{own}} < \sigma$ (Section 6 uses $\sigma_{\text{own}} = 0.1$ and broadcast noise $\sigma = \sigma_{\text{obs}} = 0.3$). No robot ever sees the clean stream, and no two robots see the same stream. We take this **persistent** mask as the primary case: an i.i.d., per-round mask reduces to standard uniform sampling, so the persistent (structured) mask is the harder regime where our recovery condition (Theorem 2) applies; the released suite supports both, and Appendix F confirms the headline results are unchanged under the i.i.d. mask.

This channel is the formal counterpart of physical sensing: a robot perceives a teammate's engagement and its outcome only when the teammate is within range (the persistent partial mask) and reads it with a fidelity that degrades with distance and with its own sensor (the per-observer noise), so it is physically realizable rather than a convenient abstraction: onboard, range-limited mutual perception without communication already drives real drone swarms [4,5]. Concretely, a robot can often assess a teammate's outcome by direct observation (for example, seeing whether a surveilled area was in fact covered, or whether a serviced target stopped emitting) without any message from the teammate; where only the teammate's action (which task it engaged), and not its scalar outcome, is observable, an action/choice channel is the natural fallback, which we develop in a follow-up. It is strictly weaker than the shared, clean broadcast usually assumed, and it makes decentralization **real**: persistent blind spots give every robot a permanently different view, and the private per-observer noise means even commonly-visible outcomes are read differently by each robot, so there is no shared, clean signal to average toward agreement and the robots cannot converge to a common model by symmetry. Figure 1 illustrates the setting.

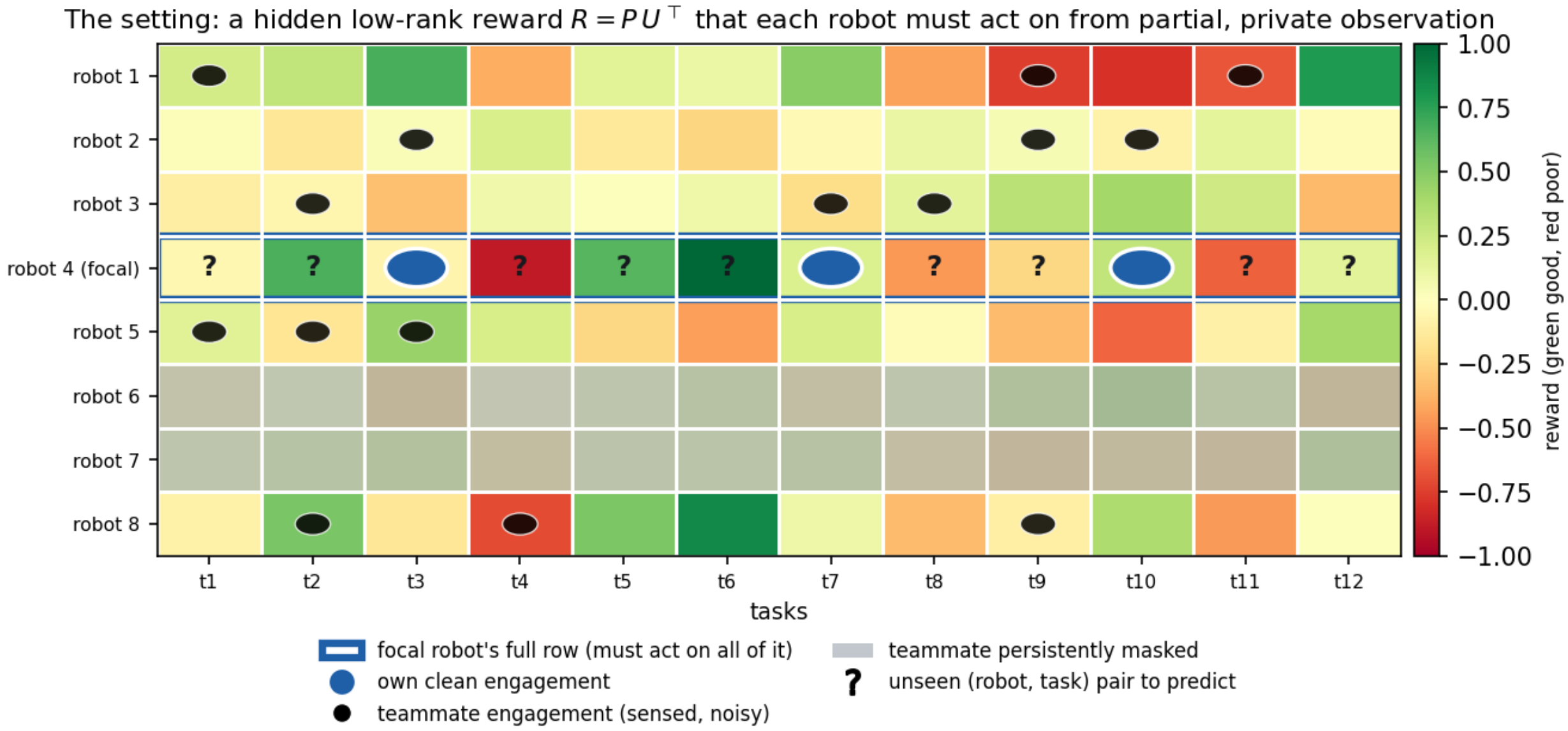

***Figure 1.*** *The setting. The robot × task reward is hidden and low-rank, $R = PU^\top$ (capability traits $p_i$, requirement traits $u_j$). A focal robot (blue row) must act on its whole row, including the many pairs it never engaged ('?'), using only its own clean engagements (blue circles), a partial and per-observer-noisy view of the teammates it can sense (black dots; greyed rows are persistently invisible to it), and no communication.*

**Objective and metric.** We measure decision quality by the normalized **skill** (a Murphy skill score [51] / normalized return),

$$\text{skill} = \frac{\text{earned} - \text{random}}{\text{oracle} - \text{random}},$$

where for an offered set the *oracle* picks $\text{argmax}_{j\in S} R_{ij}$ and *random* picks uniformly; skill $= 0$ is the no-information floor and 1 is omniscient (a policy worse than random scores below 0). We report **unseen-pair skill** (restricted to tasks the robot never engaged, the generalization test), and the **anytime** (cumulative-reward) skill over the mission.

## 4. Method: SwarmCF

Each robot independently maintains low-rank factor estimates $\hat{P} \in \mathbb{R}^{m\times\hat{d}}$, $\hat{U} \in \mathbb{R}^{n\times\hat{d}}$ and updates them online from whatever it senses, using **noise-weighted alternating least squares** (weighted ridge ALS [40,44], a standard estimator for low-rank completion) on the observed entries. The key design choice for masking-robustness: an unobserved entry receives zero *weight*, never an imputed zero *value*. The robot then acts greedily (with a small $\varepsilon$ for exploration) on its **completed** reward row, which is defined for every task, including ones it never engaged.

Algorithm 1: SwarmCF (run independently by each robot $i$)
init $\hat{P}, \hat{U}$ small-random; observed set $\Omega_i \leftarrow \emptyset$
for round $t = 1, \ldots, T$:
  offered $S_{it}$; act $a_{it} \leftarrow \text{argmax}_{j\in S_{it}} \langle \hat{p}_i, \hat{u}_j \rangle$ (w.p. $1 - \varepsilon$, else random); engage, earn $R_{i,a_{it}}$
  sense broadcast: for each visible teammate $k$ (i.e. $M_{ik} = 1$) record
    $(k,\ a_{kt},\ \tilde{r} = R_{k,a_{kt}} + \eta_{ikt})$ with weight $w = 1/\sigma^2$ (uniform $w = 1$ in our runs) into $\Omega_i$
    (own outcome recorded with its own, lower, noise)
  every $\tau$ rounds: refit by weighted ridge ALS sweeps over $\Omega_i$:

$\hat{u}_j \leftarrow (\sum_{(k,j)\in\Omega_i} w\ \hat{p}_k \hat{p}_k^\top + \lambda I)^{-1} \sum w\ \tilde{r}\ \hat{p}_k$; symmetric update for $\hat{p}_i$
predict full row $\hat{R}_{i\cdot} = \hat{U}\hat{p}_i$ (defined on EVERY task, seen or not)

**New-task onboarding (fold-in [52]).** Because the swarm already holds the robot-factor basis, a new task $j^\star$ is absorbed without retraining: its hidden vector is the ridge least-squares solution of its few observed engagements against the corresponding known robot factors, an $O(\hat{d}^3)$ computation, after which every robot can score $j^\star$. The same fold-in lets a robot predict any unseen pair once it has recovered the basis (Algorithm 2). **A new robot is different:** it arrives with no memory, and with no communication it cannot be handed the basis, so it must first recover the task factors from the passive broadcast (the coverage time of Theorem 2) and only then fold in its own $\geq \hat{d}$ engagements; its onboarding is bounded by recovery, not by the $O(\hat{d})$ fold-in.

Algorithm 2: Fold-in (solve a new entity's $\hat{d}$-vector from a few observations against a known basis $B$)
given basis $B$ (the $\hat{d}$-dim factors of the entities the newcomer has been observed against)
and observations $y$ (the few rewards seen for the newcomer), with weights $W$:
$\hat{x} \leftarrow (B^\top W B + \lambda I)^{-1} B^\top W y$ // $O(\hat{d}^3)$ ridge solve
predict reward on any other entity with factor $b$ as $\langle b, \hat{x} \rangle$

**What SwarmCF does and does not assume.** It is fully decentralized (one estimator per robot, no shared state), communication-free (it only reads the passive stream), and prior-free beyond a guessed rank $\hat{d}$ (which it does not need to be exact, and which can itself be removed by a rank-adaptive variant we defer to follow-up work). It does not assume the noise level is known: uniform weighting suffices and is what we use for the headline results.

**Computational cost.** SwarmCF is light enough to run on each robot. Acting is $O(c\hat{d})$ per round (score the offered set). A periodic refit is a few alternating ridge sweeps, each solving $O(m+n)$ linear systems of size $\hat{d} \times \hat{d}$, i.e. $O(\text{sweeps} \cdot (m+n)\hat{d}^3 + |\Omega_i|\hat{d}^2)$ time, with $O((m+n)\hat{d} + |\Omega_i|)$ memory; since $\hat{d}$ is a small guessed rank this is small for swarm-scale $m, n$. Folding a new task (or an unseen pair) into the recovered basis is a single $O(\hat{d}^3)$ ridge solve. There is no inter-robot computation: each robot updates only its own factors and reads the passive stream.

## 5. Theory: the categorical separation from structure-free learning

We formalize the separation between structure-free learning and SwarmCF, give the per-robot sample complexity, and state the conditions under which decentralized recovery from the masked broadcast succeeds. Proofs are in Appendix A; here we give the statements and the intuition. We use the term *Proposition* for a self-contained structural fact, *Lemma* for an intermediate result that a theorem builds on, and *Theorem* for a main separation or recovery guarantee. A learner is **structure-free** if its estimate of $R_{ij}$ depends only on robot $i$'s own past engagements of task $j$ and equals a fixed prior on any task it never engaged (the per-arm class: Independent-UCB, tabular).

**Proposition 1 (structure-free floor).** For a structure-free learner and any task $j$ that robot $i$ never engaged, the estimate is the prior constant, so its expected unseen-pair skill is exactly 0 and its squared error is at least the row variance $\Omega(1)$ (under the bounded, normalized reward of Section 3). Moreover the broadcast is provably uninformative to it: its per-task estimate is by definition not a function of any other task or robot.

**Lemma 1 (CF row completion, $\Theta(d)$ versus $\Theta(n)$).** If the task factors $U$ are known (rank $d$) and robot $i$ observes its true rewards on any set $\Omega$ with $|\Omega| \geq d$ whose factors span $\mathbb{R}^d$, then $p_i$ is the unique least-squares solution and $R_{ij} = \langle p_i, u_j \rangle$ is recovered **exactly for all** $j$. Per-robot sample complexity is therefore $\Theta(d)$, versus $\Theta(n)$ for any structure-free learner.

**Theorem 1 (anytime separation under task scarcity).** Offers are uniform random size-$c$ subsets of the $n$ tasks. **(a)** Every structure-free learner has expected anytime (cumulative-reward) skill at most $c(T-1)/(2n) \leq cT/(2n)$, which is $o(1)$ whenever $cT = o(n)$, even with a full broadcast. **(b)** Once the shared basis is recovered (Lemma 1, available per Theorem 2 after $T_{\text{rec}} = O((nd/\rho m)\log n)$ rounds), SwarmCF plays the recovered-row greedy policy and earns per-round skill 1 in the noiseless case (and $1 - O(\sigma\sqrt{d/|E_i(j)|})$ under noise), so for $T = \omega(T_{\text{rec}})$ its anytime skill is $\Omega(1)$. Hence whenever the team is large enough to recover within the scarce-offer horizon ($\rho m = \omega(cd\log n)$, giving a non-empty window $\omega(nd/\rho m \log n) = T = o(n/c)$) the separation is categorical: structure-free anytime skill $\to 0$ while SwarmCF stays $\Omega(1)$.

Proposition 1, Lemma 1, and Theorem 1 make the separation categorical (zero versus nonzero) and operational (it shows up in reward earned while learning). The zero side is **by construction**: a structure-free learner is *defined* to fall back on the prior off its own engagements, so its floor on unseen pairs characterizes that class (independent per-task bandits, tabular value tables) rather than reflecting a contest; the weight of the claim rests on the nonzero side, that SwarmCF can in fact recover and act, which is not automatic. Since Lemma 1 still assumes $U$ is known, the remaining question, the central problem of the decentralized setting, is whether each robot can **recover** the shared structure from its own privately-masked, noisy stream. Because the mask is over robot pairs, robot $i$'s observations form a structured (non-uniform) sub-sample, exactly where off-the-shelf uniform-sampling completion does not apply. We give a deterministic condition instead.

**Theorem 2 (decentralized masked recovery).** Let $E_i(j)$ be the set of robot $i$'s visible teammates that engaged task $j$, with factor matrix $B = P_{E_i(j)}$, and suppose $i$'s observation graph contains a $\hat{d} \times \hat{d}$ fully-observed invertible anchor block (fixing the global factor frame, i.e. the rotation gauge). Then robot $i$ predicts the pair $(i, j)$ exactly (noiseless) **iff** $p_i \in \text{span}\{p_k : k \in E_i(j)\}$; the full task vector $u_j$ is recovered when those factors span $\mathbb{R}^d$. Under per-observation noise the prediction error is $O(\sigma\sqrt{d}/\sigma_{\min}(B))$, where $\sigma_{\min}(B)$ is the smallest singular value of the engager factor matrix $B$; since $\sigma_{\min}(B)$ grows with the number of engagers (generically $\Theta(\sqrt{|E_i(j)|})$) the error decreases as $O(\sigma\sqrt{d/|E_i(j)|})$. Conversely, if $p_i \notin \text{span}\{p_k : k \in E_i(j)\}$ the pair is non-identifiable and the learner is at the prior floor. Under non-adaptive exploration the condition holds for all tasks, with high probability, after $T = O(nd/\rho m \log n)$ rounds, a rate that improves as the team grows.

Theorem 2 turns the previously-cited completion step into a self-contained, checkable condition: a pair $(i, j)$ is predicted exactly when robot $i$'s own factor lies in the span of the visible teammates that engaged $j$, and not otherwise, with the threshold reached faster the larger the team. The coverage condition needs $\rho m > d$ on average, so at very low broadcast ($\rho m < d$, e.g. $\rho = 0.1$ with $m = 30, d = 5$) recovery is only partial and the low-$\rho$ skill in Figure 2 is correspondingly reduced. Appendix C validates it directly: on the actual observation patterns the swarm produces, reconstruction error collapses from the prior floor to (numerically) zero exactly at the spanning threshold.

**Theorem 3 (collective speedup, why a swarm). (a) Necessity.** An isolated robot ($\rho = 0$) observes only its own row; for rank $d > 1$ a single row does not identify the column space, so its unseen-pair skill stays at the floor of Proposition 1 and sharing is **necessary**. **(b) Speedup.** For $\rho > 0$ the recovery time of Theorem 2 is $T_{\text{rec}} = O(nd/\rho m \log n)$, which decreases as $1/m$ as the team grows (at constant broadcast $\rho = \Theta(1)$, $T_{\text{rec}} = O((nd/m)\log n)$); a lone learner never recovers. The broadcast makes recovery possible and a larger team makes it proportionally faster, with no communication.
Theorems 2-3 are, to our knowledge, the first results that pin decentralized low-rank recovery from a persistent, private, per-robot mask to an explicit condition and tie its rate to team size; they are what make the categorical claim self-contained rather than imported from centralized theory. Full proofs are given in Appendix A.
**Relation to the experiments.** Theorems 1-3 are worst-case *sufficient* guarantees: they certify *when* recovery and the categorical gap are achievable and how the rate improves with team size, and Appendix C confirms the governing spanning condition directly on the swarm's actual coverage. SwarmCF then realizes the advantage **sample-efficiently**: acting well requires only correct *ranking* of each offered set, not exact reconstruction of the entire row, and recovery is *graded* in the local spanning rank (Appendix C), so a robot scores a task accurately as soon as a few similar teammates have engaged it. The headline horizon $T = 50$ is therefore the natural operating point: the theory establishes feasibility and the team-size scaling, and the experiments show the method reaches strong skill comfortably within that envelope.

## 6. Experiments

**Setup.** Unless noted, $m = 30$ robots, $n = 240$ tasks, true rank $d = 5$, guessed rank $\hat{d}$ drawn at random in $[d, 2d]$ per run (no method is given the true rank; robust to the guess, Figure 7), horizon $T = 50$, partial broadcast $\rho$ swept, private noise on own ($\sigma_{\text{own}} = 0.1$) and observed ($\sigma_{\text{obs}} = 0.3$) outcomes (Appendix D), 16 random seeds, bootstrap 95% confidence intervals. A robot may in principle select any of the $n$ tasks; the body sweeps restrict each offer to a uniform random size-$c$ subset ($c = 20$) to control scarcity and to supply the per-round engagement diversity the online estimator relies on. This is a moderate scarcity (each task is offered about $cT/n \approx 4$ times); Theorem 1's strict scarce-offer regime ($cT = o(n)$) is shown separately in Appendix F (Figure 11a). Appendix F reports the unrestricted all-tasks menu ($c = n$) as a robustness check: the categorical low-rank separation is unchanged, while the online method's lead over batch completion narrows. Because no existing benchmark instantiates the ZK-MRTA regime (a hidden low-rank robot-task reward seen only through a persistent, per-observer-private, masked and noisy broadcast under communication-free, task-scarce decentralized choice), we built and openly released the **LatentSwarm** simulator and run every experiment on it (Section 6.5 adds capacity-1 contention; Appendix E details the simulator).
**How to read the comparison.** The setting itself is new, so this is a controlled sweep across the low-rank design space against the genuinely external structure-free paradigm and full-information reference ceilings, not a contest of rival systems. SwarmCF is our method; structure-free learners (Independent-UCB, a per-arm UCB1 [53]; tabular) are the external paradigm; standard low-rank estimators (online MF-SGD, batch spectral and Bayesian completion) are adapted to the setting for the low-rank comparison; the Oracle and a centralized full-communication matcher are upper bounds, not competitors. We emphasize that communication-based methods (auctions/CBBA, CTDE training, federated/gossip CF) are **inadmissible by the problem definition**, not omitted: they require messages or a coordinator, which our setting forbids, so they can appear only as the centralized ceilings. The admissible communication-free competitors are exactly the structure-free learners (Independent-UCB / tabular), which are the per-arm reductions of no-communication multiplayer-bandit methods. In our harness **every** method runs decentralized and communication-free (one estimator per robot); the low-rank methods differ only in the update rule. Table 2 fixes each method's operating profile.

*Table 2. Operating profiles of the methods compared; SwarmCF-family refinements are deferred to future work (column abbreviations are defined in the key).*

**Column key. Observability:** full = every engagement seen noiselessly; ρ = a fraction ρ of engagements seen (masked); ρσ = masked and read with per-observer noise; fullσ = unmasked but noisy; self = own outcomes only. **Prior:** – = none; $\hat{d}$ = a guessed rank; d = the true rank; U* = oracle factors. **Computation:** ETC = explore-then-commit. In our harness every method runs decentralized and communication-free (one estimator per robot on the passive broadcast); the low-rank methods differ only in the update rule, so the comparison is controlled. **SwarmCF** (our online method) and its batch variant **SwarmCF-batch** are ours (shaded); the rest are standard low-rank estimators we adapt, the structure-free paradigm, or reference ceilings (full communication, not competitors).

| Method | Provenance | Distribution | Communication | Observability | Prior | Computation |
|---|---|---|---|---|---|---|
| **SwarmCF** | ours | decentralized | none | ρσ | $\hat{d}$ | online |
| **SwarmCF-batch** | ours | decentralized | none | ρσ | $\hat{d}$ | batch |
| MF-SGD | standard, adapted | decentralized | none | ρσ | $\hat{d}$ | online |
| BiasModel | standard, adapted | decentralized | none | ρσ | – | online |
| BPMF | standard, adapted | decentralized | none | ρσ | $\hat{d}$ | batch |
| CLUB | standard, adapted | decentralized | none | ρσ | – | batch |
| ESTR | standard, adapted | decentralized | none | ρ | $\hat{d}$ | ETC |
| Independent-UCB | structure-free baseline | decentralized | none | ρσ | – | online |
| Tabular | structure-free baseline | decentralized | none | ρσ | – | online |
| Random | structure-free baseline | decentralized | none | – | – | – |
| Centralized (clean) | reference (upper bound) | centralized | full | full | $\hat{d}$ | online |
| Centralized (noisy) | reference (upper bound) | centralized | full | fullσ | $\hat{d}$ | online |
| Oracle | reference (upper bound) | centralized | full | full | U* | – |

## 6.1 The categorical separation and masking robustness

Figure 2 sweeps the broadcast rate $\rho$ and reports unseen-pair skill. SwarmCF acts well on tasks it never engaged at every broadcast rate, while the structure-free learners sit at the floor ($\approx 0$) by construction, the categorical separation of Proposition 1 and Lemma 1. Among low-rank methods, SwarmCF's online updates stay robust as the broadcast is masked, whereas the batch variant SwarmCF-batch (which imputes unobserved entries) decays; the two cross over near $\rho \approx 0.6$ and it leads only at full broadcast, where its one-shot factorization on a near-complete matrix is the best case for completion. The operationally relevant regime, partial observation, is exactly where the online estimator leads.

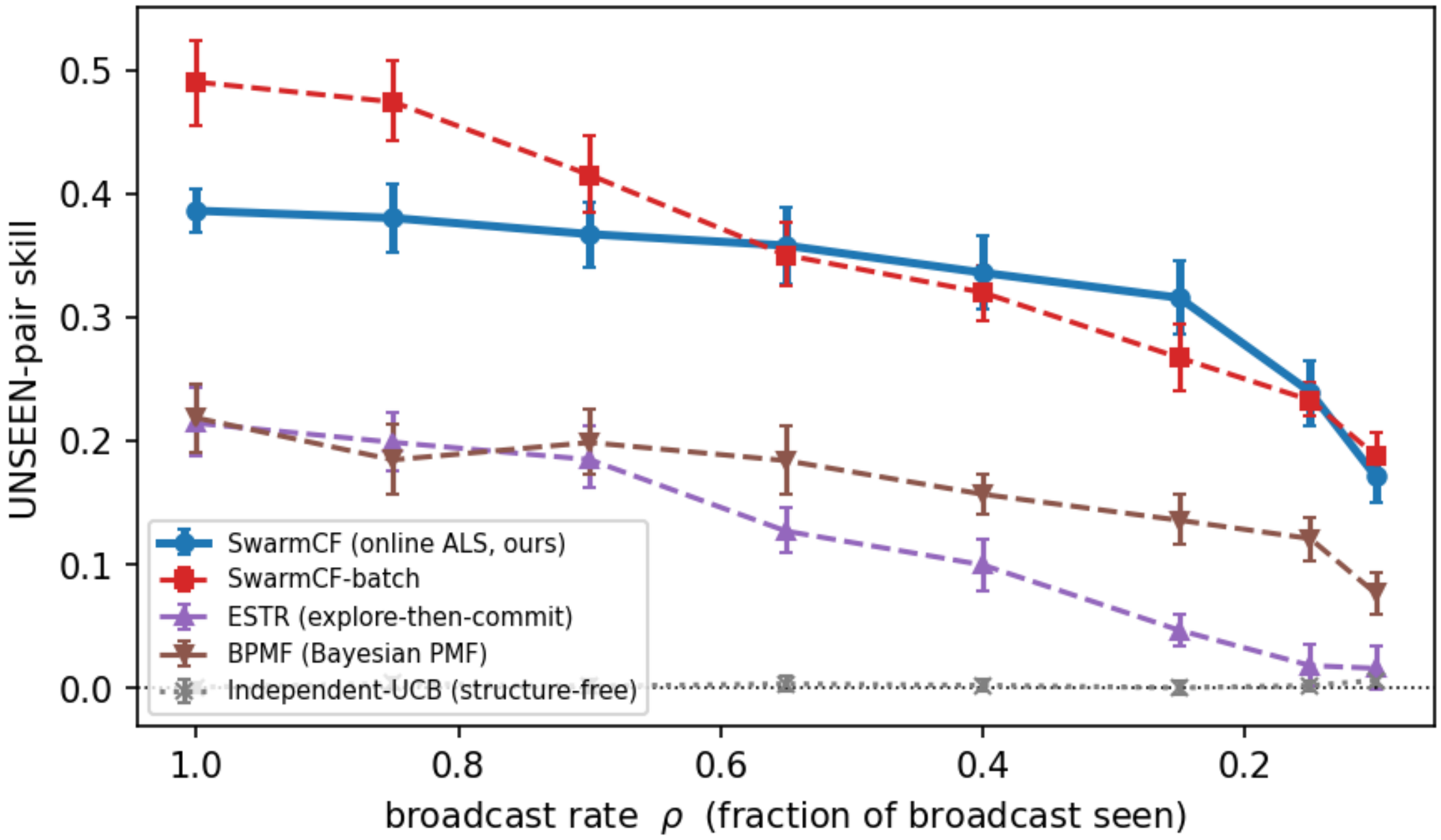

***Figure 2.*** *Unseen-pair skill versus broadcast rate ρ. Structure-free learners are pinned at the floor at every broadcast rate; SwarmCF acts on the unseen throughout and is robust under masking, while the batch variant SwarmCF-batch decays as observation becomes partial. Means with bootstrap 95% CIs over 16 seeds.*

### 6.2 The operational (anytime) separation

Final-policy quality can overstate a method that explores cheaply. The operationally relevant measure is reward *earned while learning*. Figure 3 shows cumulative-reward skill over the mission: SwarmCF earns from the first rounds. Independent-UCB stays near the random floor: with $n \gg T$ arms its optimism keeps it exploring untried tasks. Phase-structured low-rank methods pay an explore-then-commit penalty early. An $\varepsilon$-greedy tabular learner does earn some reward by re-exploiting tasks it has already engaged (each task recurs in offers about $cT/n \approx 4$ times here), but it stays well below SwarmCF and at the floor on unseen pairs (Table 3). The clean anytime collapse of Theorem 1 holds in the strict regime $cT = o(n)$ (demonstrated in Appendix F, Figure 11a); SwarmCF's early-earning advantage is broader, as seen here.

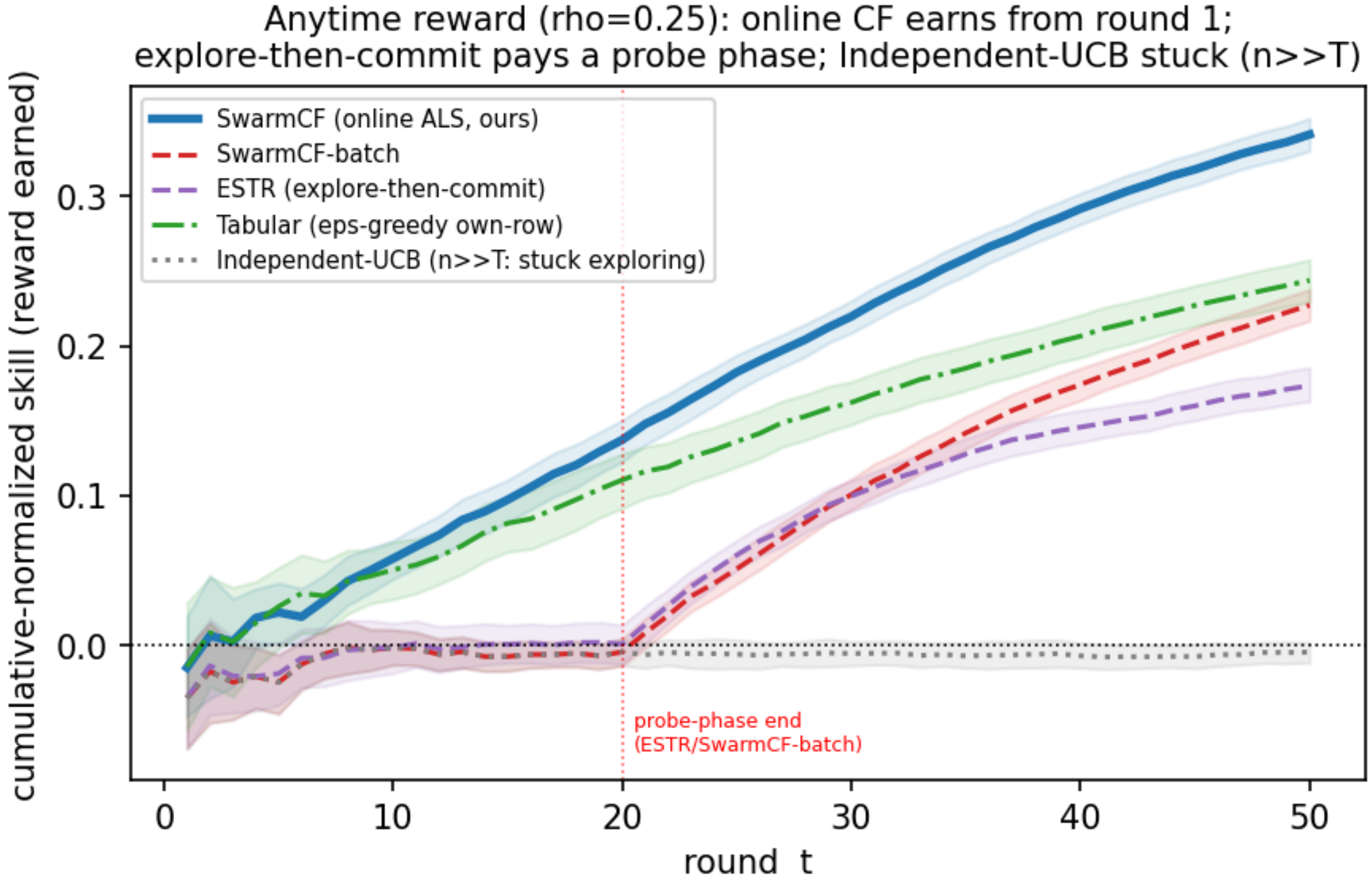

***Figure 3.*** *Anytime cumulative-reward skill ($\rho = 0.25$). SwarmCF earns from round one; explore-then-commit pays a probe phase; Independent-UCB stays near the random floor, while $\varepsilon$-greedy tabular earns only by re-exploiting tasks it has already engaged. Means with bootstrap 95% CIs over 16 seeds.*

### 6.3 Why a swarm: the value of the broadcast and a positive scaling law

Two experiments isolate what the team and the broadcast contribute (Figure 4). *(a) Value of the broadcast.* Sweeping from $\rho = 0$ (each robot isolated, sees only its own outcomes) to $\rho = 1$ (full passive sensing): a lone robot cannot recover the shared structure from its single matrix row, so isolated unseen skill is $\approx 0$; the broadcast lifts SwarmCF by $+0.39$ unseen skill but a structure-free learner by $\approx 0$, which has no model linking tasks and so cannot use it (Theorem 3). *(b) Positive scaling.* Holding $n$, horizon and $\rho$ fixed and growing the team from $m = 5$ to 80, SwarmCF's unseen skill rises monotonically ($0.13 \rightarrow 0.43$): more robots feed more observations into the one shared structure, so each robot's competence on tasks it never engaged grows with the team. The batch variant SwarmCF-batch rises even more steeply (Figure 4b), overtaking the online variant for large teams as the pooled observations sharpen its one-shot refit (mirroring the broadcast-rate crossover of Figure 2), so the positive scaling is a structural property of the shared low-rank reward, not specific to the online update. Structure-free learning is flat. A swarm whose per-robot competence rises with team size, the opposite of the usual interference penalty, is a direct consequence of sharing structure.

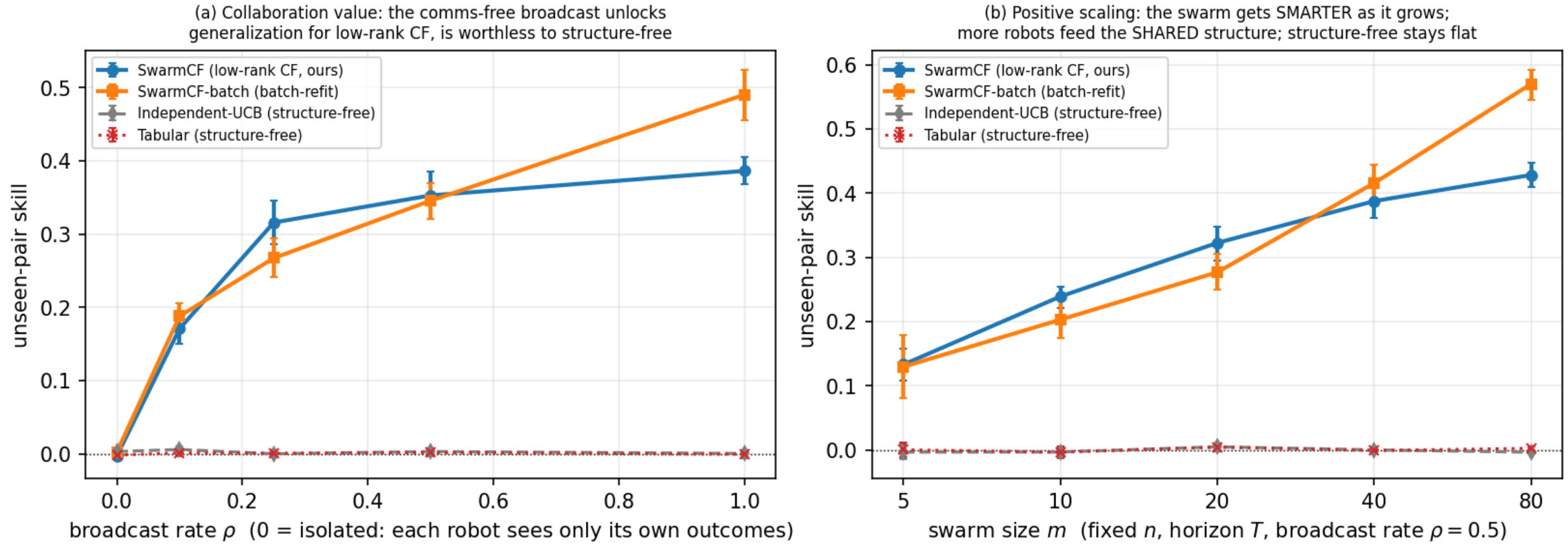

***Figure 4.*** *(a) Value of the broadcast: unseen skill versus $\rho$ (left edge = isolated). (b) Positive scaling: unseen skill versus team size $m$ at fixed broadcast rate $\rho = 0.5$ (and fixed $n$, horizon $T$). In both panels our online SwarmCF and its batch variant SwarmCF-batch rise (the gain is structural), while structure-free learners stay flat. Means with bootstrap 95% CIs over 16 seeds.*

## 6.4 The cost of communication-free operation: a centralized ceiling

Table 3 consolidates the masked-harness comparison across all methods, including two structured controls that bound the explanation of the advantage. A published **clustering-of-bandits** baseline (CLUB [49], which shares structure through discrete robot clusters rather than continuous factors) also generalizes well above the structure-free floor (unseen-pair skill 0.26 at $\rho = 0.25$), so the categorical gap is *structure-sharing versus none* rather than an artifact of our particular estimator, though SwarmCF's continuous low-rank still leads it under masking (0.32 versus 0.26); an additive **popularity** model (BiasModel [41], rank $\leq 2$) captures only globally-good tasks and reaches just 0.09, so the advantage is *personalized* low-rank transfer, not a shared popularity ranking. Here we ask how far communication-free SwarmCF is from a centralized optimum that our constraints forbid. We add two reference **ceilings** (upper bounds, not competitors): a centralized full-communication matcher that sees every reward and computes the optimal one-to-one robot-task assignment by Hungarian matching [54], evaluated both with clean (noiseless) observation and under the realistic per-observer noise. Under limited observability ($\rho = 0.25$, matching-normalized anytime earned skill at the headline $n = 240$) SwarmCF earns 0.42 against the clean full-communication ceiling's 0.53, recovering about 80% of it (Table 4); the same matcher under per-observer noise reaches a statistically indistinguishable 0.54, so observation noise costs the centralized matcher essentially nothing (the optimal assignment depends only on the ordering of rewards, which zero-mean observation noise rarely changes once reward gaps exceed the noise scale), and the price of the masked, privately-noisy observation is small *on this operational metric*: on earned (anytime) skill the estimator is **nearly at the centralized ceiling** (the harder unseen-pair generalization of Table 3 is naturally lower), so the residual gap to the centralized optimum is within-round **coordination**, not estimation. Coordination is therefore the binding constraint, which Section 6.5 isolates by adding capacity-1 contention.

***Table 3.*** *Performance scorecard on one canonical masked harness.*

| Method | Provenance | unseen skill (ρ=0.25, masked) | unseen skill (ρ=1, full) | cumulative regret (ρ=0.25, lower=better) | rounds to 25% of oracle |
|---|---|---|---|---|---|
| **SwarmCF** | ours | 0.316 [0.286, 0.346] | 0.386 [0.369, 0.404] | 41.1 | 34 |

| Method | Provenance | unseen skill (ρ=0.25, masked) | unseen skill (ρ=1, full) | cumulative regret (ρ=0.25, lower=better) | rounds to 25% of oracle |
|---|---|---|---|---|---|
| **SwarmCF-batch** | ours | 0.267 [0.241, 0.294] | 0.490 [0.455, 0.524] | 46.2 | never |
| MF-SGD | standard, adapted | 0.007 [-0.006, 0.019] | 0.048 [0.035, 0.060] | 46.7 | never |
| BPMF | standard, adapted | 0.136 [0.117, 0.157] | 0.219 [0.191, 0.246] | 49.7 | never |
| ESTR | standard, adapted | 0.047 [0.034, 0.059] | 0.214 [0.189, 0.243] | 46.6 | never |
| CLUB | standard, adapted | 0.258 [0.232, 0.284] | 0.436 [0.413, 0.463] | – | – |
| BiasModel | standard, adapted | 0.094 [0.056, 0.134] | 0.131 [0.091, 0.172] | – | – |
| Independent-UCB | structure-free baseline | 0.000 [-0.004, 0.005] | 0.001 [-0.004, 0.005] | 50.4 | never |
| Tabular | structure-free baseline | 0.001 [-0.004, 0.006] | -0.000 [-0.005, 0.004] | 43.4 | 46 |
| Random | structure-free baseline | 0.001 [-0.012, 0.013] | -0.003 [-0.018, 0.012] | 50.1 | never |

One canonical masked harness (m=30, n=240, 16 seeds); unseen-skill columns report the mean with a bootstrap 95% confidence interval in brackets, regret and time-to-competence are means from the anytime trajectories. SwarmCF leads the operational columns (lowest regret, fastest to competence) and is the most masking-robust; on masked unseen skill its margin over the batch variant SwarmCF-batch is within the 16-seed interval, and the batch and clustering baselines win only the full-broadcast column. Structure-free learners are at the floor on the unseen columns (intervals straddling zero); on the operational columns an ε-greedy tabular learner is competitive by re-exploiting already-engaged tasks, so the categorical separation is specifically an unseen-pair (generalization) phenomenon.

***Table 4.*** *The cost of communication-free operation: earned (anytime) skill against the centralized full-communication Hungarian ceiling with clean observation (matching-normalized, $\rho = 0.25$, $n = 240$, means with bootstrap 95% CIs over 16 seeds). SwarmCF recovers about 80% of the ceiling with no communication, while the structure-free learner is near the floor; the same matcher under realistic per-observer noise reaches a statistically indistinguishable 0.54, so observation noise costs the centralized matcher essentially nothing.*

| Method | earned skill (ρ=0.25, anytime) | fraction of the ceiling |
|---|---|---|
| Centralized full-communication, Hungarian (clean ceiling) | 0.528 [0.505, 0.552] | 1.00 |
| **SwarmCF** (ours, communication-free) | 0.423 [0.407, 0.441] | 0.80 |
| Independent-UCB (structure-free) | 0.003 [-0.003, 0.009] | 0.01 |

### 6.5 Capacity-1 contention and communication-free de-confliction

The experiments so far impose no contention: two robots may service the same target. We now turn on **capacity-1 contention** in LatentSwarm, where only the first robot to select a task each round succeeds and colliding robots waste the engagement, the within-round coordination cost Section 6.4 identified as the binding constraint. Nothing else changes: the same signed low-rank reward, the same persistent private mask, the same task scarcity and randomly guessed rank, with SwarmCF run as one policy under no contention handling.

**(a) The categorical separation survives contention.** SwarmCF earns the most of any learner (42% of the centralized capacity-1 ceiling, bootstrap 95% CI [40%, 44%] over 16 seeds; this capacity-1 ceiling is stricter than the full-communication ceiling of Section 6.4, so the drop from about 80% there to 42% here is the coordination gap that contention introduces), well above the MF-SGD baseline (20%) and the structure-free independent-UCB learner, which sits below the random floor (-17%) because under contention its persistent exploration collides without coordinating. On the categorical generalization metric, SwarmCF is the **only** method whose **unseen-pair** skill is significantly above the floor (0.58, CI [0.52,0.64]; MF-SGD, UCB, and random all have intervals straddling zero): the Proposition 1 separation holds under contention (Figure 5). This contention study uses $\rho = 0.5$, the all-tasks menu, and the stricter capacity-1 Hungarian normalization, so its skill values run higher than, and are not directly comparable to, the masked-harness numbers of Table 3 (which use $\rho = 0.25$ and a size-$c$ menu); only the categorical above-floor-versus-at-floor pattern is meant to carry across settings.

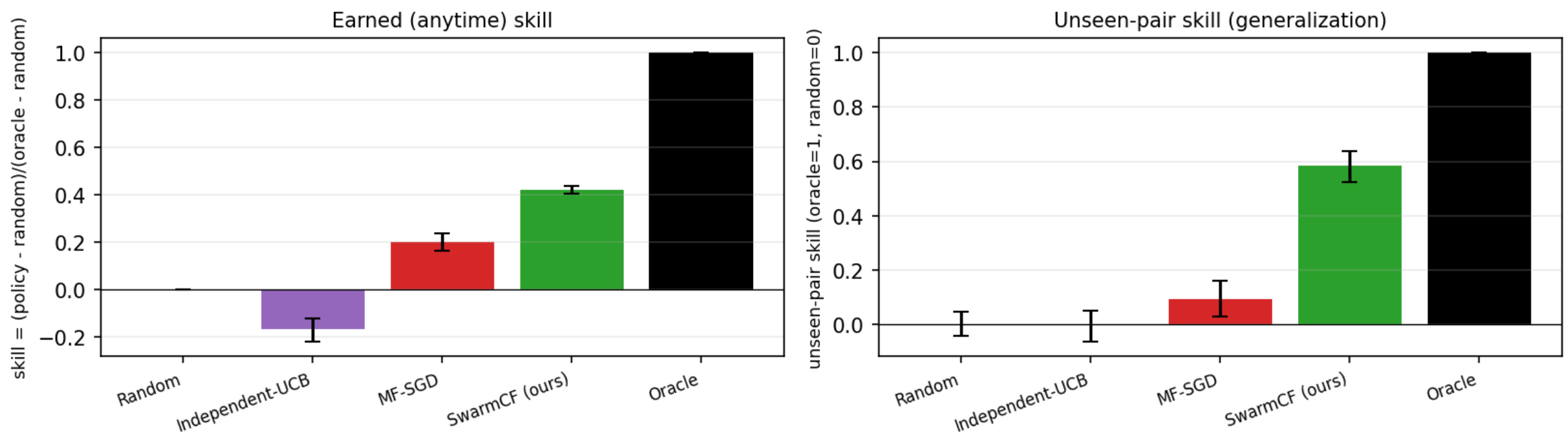

***Figure 5.*** *The categorical separation under capacity-1 contention (LatentSwarm: signed low-rank reward, persistent partial and private broadcast, task scarcity). Left: earned (anytime) skill, SwarmCF best among learners and far above the structure-free floor (independent-UCB goes negative under contention). Right: unseen-pair skill, SwarmCF the only method significantly above the floor. Bars are means with bootstrap 95% CIs over 16 seeds; the oracle is the centralized capacity-1 (Hungarian) matching.*

**(b) Why coordination is the binding constraint, and how SwarmCF already mitigates it.** Because the all-tasks menu offers every robot the same full set, greedy policies tend to converge on the same few high-value tasks and collide. The effect is stark (Figure 6): the structure-free independent-UCB learner collides on almost every engagement (collision rate $\approx 0.97$) and earns below random, whereas SwarmCF collides far less (rate $\approx 0.12$) because its *heterogeneous* learned models send different robots to different tasks, an implicit de-confliction that emerges from the personalized low-rank estimate with no message passing. Restricting the offer to a size-$c$ menu reduces collisions for every method (independent-UCB from 0.97 to 0.25), one reason the body uses a size-$c$ menu. An explicit communication-free de-confliction mechanism would close the remaining gap to the centralized ceiling: in a follow-up, a fixed, private per-robot offset (drawn once and never shared) roughly doubles earned reward at the most severe contention with no communication, exceeding communication-free reductions of auction-with-backoff and musical-chairs re-seating.

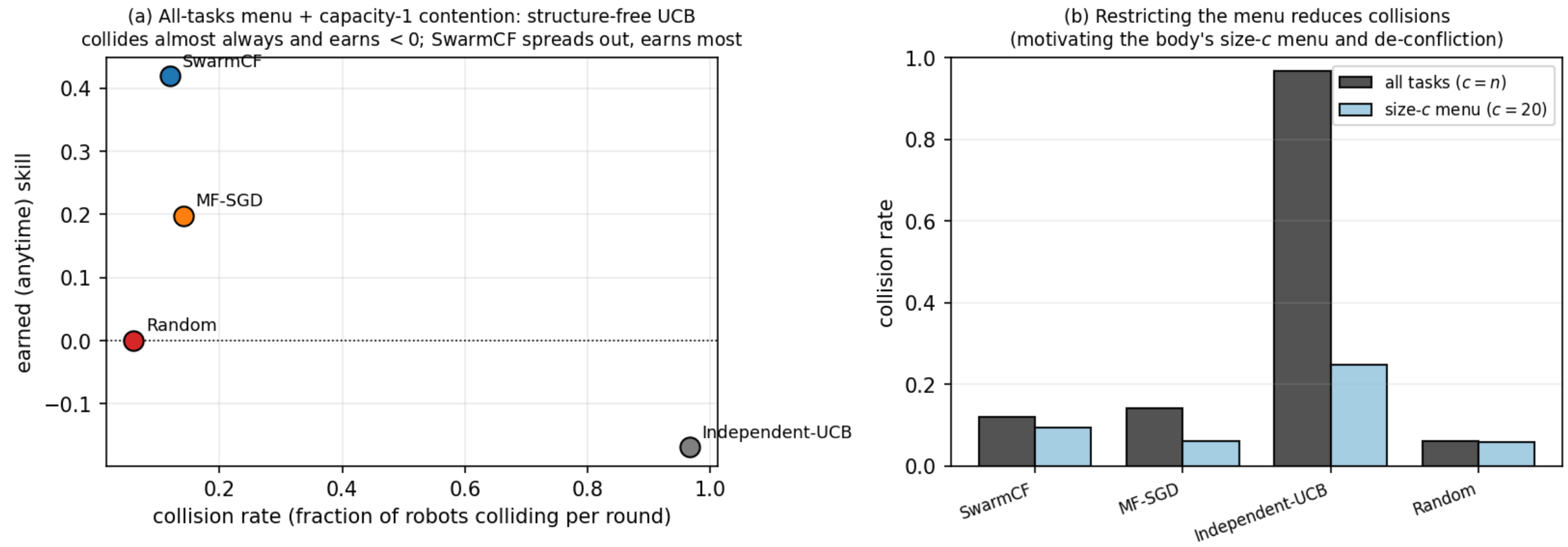

***Figure 6.*** *The cost of no de-confliction under capacity-1 contention (LatentSwarm). (a) Earned skill versus collision rate on the all-tasks menu: structure-free independent-UCB collides on almost every engagement and earns below random, while SwarmCF spreads robots across tasks via its heterogeneous learned models and earns most. (b) Collision rate per method under the all-tasks menu versus a size-$c$ menu: restricting the menu reduces collisions. Means over 16 seeds.*

### 6.6 Robustness across configurations

The separation is structural rather than tuned: it follows from the three scope conditions of Section 6.8 (an exploitable low-rank-but-personalized reward, task scarcity, and a shared channel), not from any particular team size or task count: Section 6.3 shows the separation widening with team size (Figure 4b), and additional sweeps over $n$, $d$, and reward heterogeneity in the released data show the same qualitative pattern. The categorical separation also degrades only gracefully when the reward is approximately rather than exactly low-rank (Appendix F, Figure 12). Two evaluation choices, the offer size and the masking model, are likewise robustness knobs rather than load-bearing assumptions: restricting the menu to a size-$c$ subset versus offering all $n$ tasks, and a persistent versus an i.i.d. mask, both leave the categorical separation intact (Appendix F).

**Robustness to the guessed rank.** Because the true rank is unknown, the guessed rank $\hat{d}$ is drawn at random per run; we sweep it from $d/2$ (under-ranking) to $3d$ (over-ranking) at true rank $d = 5$. SwarmCF degrades **gracefully** when the model is under-specified ($\hat{d} < d$: unseen-pair skill 0.47 at $\hat{d} = 2$, rising to 0.52 at $\hat{d} = 4$) and is **flat** once the rank is sufficient ($\hat{d} \geq d$: 0.58 to 0.64 up to $\hat{d} = 3d$), because surplus directions are shrunk by the ridge; MF-SGD stays near the floor throughout. Over-guessing is therefore safe and the exact rank is not needed (Figure 7). The guess can even be removed entirely: in a follow-up, automatic relevance determination prunes the latent directions the masked design does not excite and recovers a stable effective rank independent of the guess (whether $\hat{d} = 8$ or $\hat{d} = 20$).

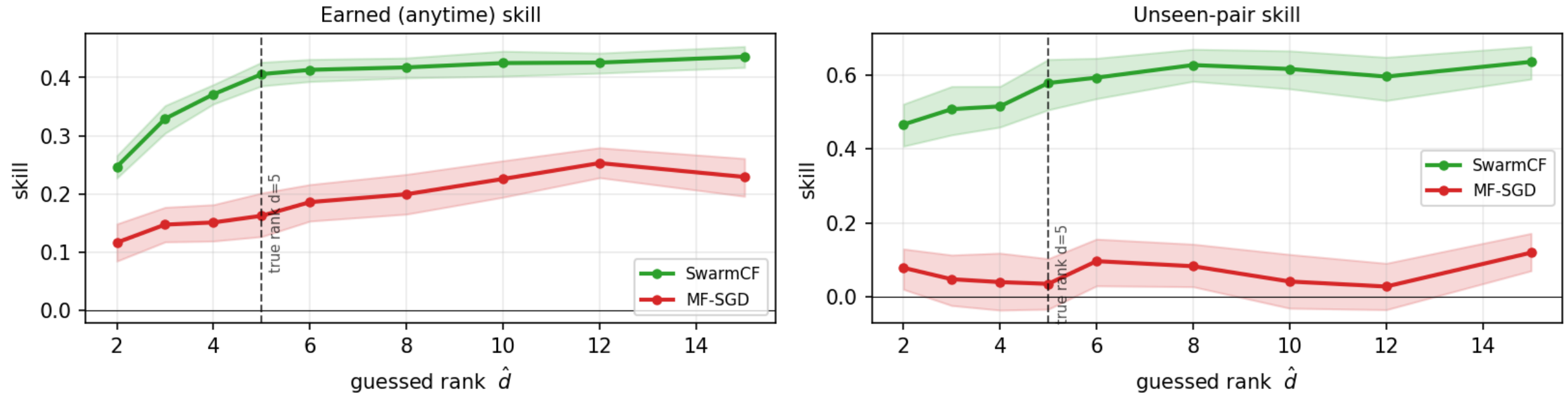

***Figure 7.*** *Robustness to the guessed rank $\hat{d}$ (true rank $d = 5$). Under-ranking ($\hat{d} < d$) is mis-specified and degrades smoothly; over-ranking ($\hat{d} > d$, up to $3d$) is regularized away and stays flat. Lines are means with bootstrap 95% CIs over 16 seeds.*

### 6.7 A robotics-grounded instance

The experiments above use abstract latent traits and a Bernoulli visibility mask. To check the separation is not an artifact of that abstraction, we instantiate the Section 3 model in a physically-grounded form. Robot capability and task requirement vectors are non-negative profiles over $d$ **sensing modalities** (electro-optical, infrared, acoustic, LiDAR, range-endurance), so the reward $R_{ij} = \langle p_i, u_j \rangle$ is a modality match and is rank-$d$ by construction (a capability-aggregation model in the style of trait-based heterogeneous MRTA [1,19,20]); and the observation mask is a **range-limited line-of-sight** graph induced by 2-D patrol positions (robot $i$ senses teammate $k$ only within a sensing radius), with per-observer noise that grows with distance ($\sigma^2 \propto 1 + (r/R_s)^2$, the free-space sensing law of coverage control [10]). Positions are thus load-bearing rather than inert, and visibility is persistent because the patrol geometry is quasi-static. Under capacity-1 contention and a randomly guessed rank, the categorical separation holds: SwarmCF reaches unseen-pair skill 0.19 (95% CI [0.15,0.24]) and earns 32% of the centralized ceiling, while structure-free Independent-UCB and MF-SGD sit at the floor (intervals straddling zero) and Independent-UCB earns below random by colliding (Figure 8). The absolute skill is lower than in the body because geometry-limited line-of-sight visibility is a harsher channel, so these values are a separate grounded instance rather than a restatement of Table 3; the separation itself is a property of the shared low-rank structure, not of the abstract trait or mask model.

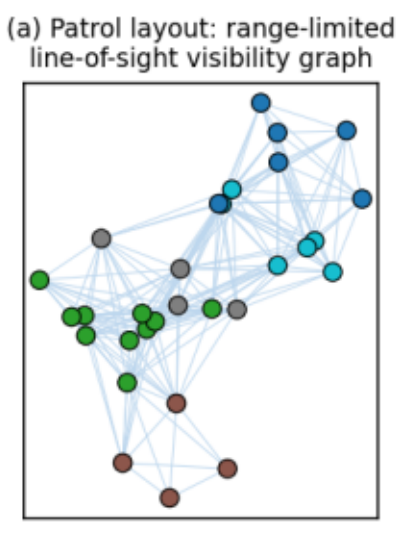

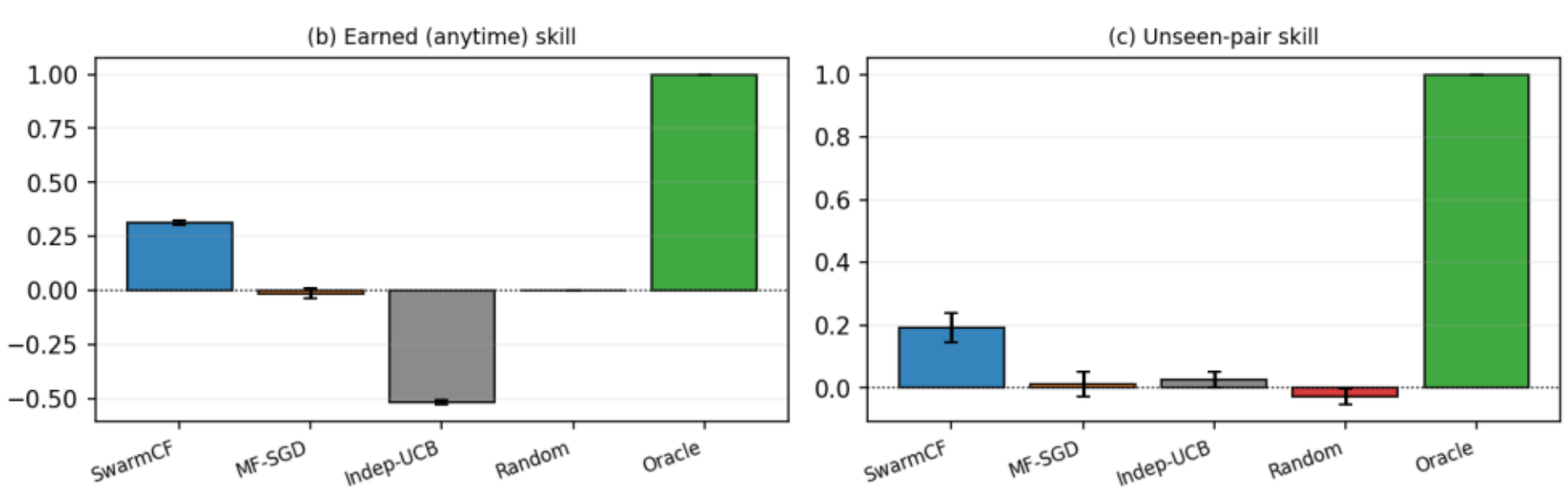

***Figure 8.*** *A robotics-grounded instance (LatentSwarm): non-negative sensing-modality traits and a geometry-induced range-limited line-of-sight mask with distance-dependent noise, under capacity-1 contention. (a) An example patrol layout and its line-of-sight visibility graph; (b) earned (anytime) skill; (c) unseen-pair skill. SwarmCF stays well above the structure-free floor, so the categorical separation survives a physically-motivated instantiation. Bars are means with bootstrap 95% CIs over 16 seeds.*

### 6.8 Scope of the advantage

The advantage is not universal, and we state its boundary precisely. It holds when three conditions co-occur: **(i) low-rank but personalized** structure ($1 < d \ll \min(m, n)$): at $d = 1$ the reward reduces to a shared popularity order that a bias/pooling baseline already captures, so there is nothing personal to transfer; **(ii) task scarcity** ($n \gg T$): if instead sample-rich, a tabular learner eventually measures every entry and the unseen advantage vanishes; **(iii) a shared channel** ($\rho > 0$): with no broadcast each robot has only its own row and collaborative filtering degenerates to tabular. These are exactly the conditions of the regime we target, and they delimit where structure-free methods remain competitive.

## 7. Discussion, limitations, and future work

**What the results say.** Under the least information (no prior, no communication, partial and privately-noisy observation), a single simple estimator gives a swarm a capability that structure-free learning provably cannot have: acting well on the unseen, getting more competent as the team grows, and, on earned (anytime) skill, recovering most of what a centralized, communicating system could achieve. The separation is structural (it is a property of exploiting the shared low-rank trait structure) and operational (it shows up in reward earned while learning, and in the operational anytime and contention metrics).

**Limitations.** The reward is assumed (approximately) low-rank and stationary, the standard trait-based premise; the categorical advantage degrades gracefully as the structure becomes only approximately low-rank or the rank grows toward full, vanishing only when there is no exploitable structure (Appendix F, Figure 12). All our evidence is in simulation: as no existing benchmark captures this regime (Section 6) we evaluate on our own released LatentSwarm simulator, and the low-rank premise, though standard for trait-based MRTA, is an assumption rather than a measured property of any specific deployment; building a higher-fidelity or physical instantiation of the regime is itself an open problem. Rewards are real-valued and bilinear in latent traits; and the recovery rate of Theorem 2 is established for non-adaptive exploration, the finite-time rate under a strongly exploiting policy (which can starve low-reward tasks of coverage) remains open. The scope conditions for the advantage (low-rank but personalized structure, task scarcity, and a shared channel) are stated precisely in Section 6.8; outside them, structure-free methods are competitive.

**Future work (a planned follow-up).** The present paper deliberately keeps to a single core estimator. In a follow-up we plan to study the refinements that this foundation enables, each of which we have prototyped: *(i)* confidence-directed exploration via a Bayesian posterior over the factors (collective, information-directed probing through the shared broadcast); *(ii)* communication-free de-confliction under capacity-1 contention via a fixed private offset, against no-communication auction and musical-chairs primitives; *(iii)* rank self-determination (removing the guessed $\hat{d}$); *(iv)* the action/choice channel as a noise-immune alternative to cardinal rewards; and *(v)* non-stationarity and team churn. We also plan a tightening of the adaptive-policy coverage rate.

## 8. Conclusion

We formalized multi-robot task allocation in its most restrictive but practically common form (no prior knowledge, no communication, partial and privately-noisy observation), and showed that decentralized online collaborative filtering over the passive broadcast lets each robot act well on tasks it has never attempted. The advantage over structure-free learning is categorical and proven; the broadcast is what makes it possible and the team is what makes it fast; and the method recovers most of a centralized full-communication ceiling on earned (anytime) skill while assuming far less. We hope the setting, a swarm that must learn to coordinate from only what it passively senses, becomes a useful baseline regime for autonomous multi-robot systems.

## CRediT authorship contribution statement

**Alexander Apartsin:** Conceptualization, Methodology, Formal analysis, Visualization, Writing - original draft. **Yigal Meshulam:** Software, Validation, Investigation. **Yehudit Aperstein:** Conceptualization, Supervision, Writing - review & editing.

## Declaration of competing interest

The authors declare that they have no known competing financial interests or personal relationships that could have appeared to influence the work reported in this paper.

## Funding

This research did not receive any specific grant from funding agencies in the public, commercial, or not-for-profit sectors.

## Data availability

The source code, the per-seed data required to reproduce every figure and table, and the **LatentSwarm** simulator (the masked-broadcast setting and its capacity-1 contention extension) are openly available at github.com/ApartsinProjects/ZKDroneSwarm.

## Appendix A. Proofs of the main results

We give proofs of the main results below (full for Proposition 1, Lemma 1, Theorem 1, and Theorem 2; Theorem 3 follows as a corollary of Theorem 2), each closed by a short remark on the proof technique and its relation to existing results. Throughout, factors are assumed in general position (generic $P, U$), the persistent mask is $M_{ik} \sim \text{Bernoulli}(\rho)$ over robot pairs, and $\hat{d} \geq d$.

**Proposition 1 (floor).** For an unobserved $j$ the estimate is a pre-chosen constant $b$; $\mathbb{E}[(b - R_{ij})^2] = (b - \mu_i)^2 + \text{Var}_j \geq \text{Var}_j = \Omega(1)$, and on an offer of never-engaged tasks selection is independent of their rewards, giving skill 0. The broadcast cannot help a per-task estimate by definition. *Remark:* elementary; it fixes the floor against which the categorical claim is measured.

**Lemma 1 (row completion).** Stacking the observed entries gives $R_{i,\Omega} = U_\Omega p_i$; spanning makes $U_\Omega$ full column rank, so $p_i = (U_\Omega^\top U_\Omega)^{-1} U_\Omega^\top R_{i,\Omega}$ is unique and exact, hence all $R_{ij}$. *Remark:* the linear algebra is standard given $U$; the contribution is the $\Theta(d)$-versus-$\Theta(n)$ contrast against the floor of Proposition 1.

**Theorem 1 (anytime).** *(a)* A structure-free learner's estimate on any never-engaged task equals the prior, so on an offer it can only rank tasks it has already engaged and is reward-blind on every other offered task. Its per-round skill is thus $\leq 1$ when the offer contains an already-engaged task and has expectation 0 otherwise, so $\mathbb{E}[\text{skill}_t] \leq \Pr[\text{offer contains an engaged task}]$. After $t-1$ rounds at most $t-1$ distinct tasks are engaged, and each lies in the uniform size-$c$ offer with probability $c/n$, so a union bound gives this probability $\leq c(t-1)/n$. Averaging over the mission (with i.i.d. offers the oracle-minus-random normaliser is identically distributed across rounds) yields expected cumulative-reward skill $\leq 1/T \sum_{t=1}^{T} c\,(t-1)/n = c(T-1)/(2n)$. *(b)* After recovery (Theorem 2) the completed row is exact in the noiseless case, so greedy selection earns per-round skill 1 thereafter; for $T = \omega(T_{\text{rec}})$ the pre-recovery rounds are a vanishing fraction, giving anytime skill $\Omega(1)$ (under noise, $1 - O(\sigma\sqrt{d/|E_i(j)|})$ per round). *Remark:* the constant is explicit (1/2); a tighter distribution-dependent rate replaces $c(t-1)/n$ by the expected normalised best-of-subset order statistic. We also confirm the bound empirically.

**Theorem 2 (recovery).** A fully-observed invertible $\hat{d} \times \hat{d}$ block pins the factor frame; per task, $R_{E_i(j),j} = B\,u_j$ with $B = P_{E_i(j)}$ determines $u_j$ uniquely iff $B$ has full column rank, and determines the pair $\langle p_i, u_j \rangle$ iff $p_i \in \text{span}\{p_k : k \in E_i(j)\}$ (the per-task analogue of Proposition 1). With noise the error is $O(\sigma\sqrt{d}/\sigma_{\min}(B)) = O(\sigma\sqrt{d/|E_i(j)|})$ since generically $\sigma_{\min}(B) = \Theta(\sqrt{|E_i(j)|})$. Coverage time (non-adaptive policy): each visible teammate engages each task with probability $\approx 1/n$ per round (for any menu size $c$: task $j$ is offered with probability $c/n$ and then selected with probability $\approx 1/c$), so it has engaged task $j$ at least once after $T$ rounds with probability $1 - (1 - 1/n)^T \approx 1 - e^{-T/n}$; the number of distinct visible engagers of $j$ is $\text{Binomial}(|N_i|, 1 - e^{-T/n})$ with $|N_i| \approx \rho m$, and requiring $\geq d$ spanning engagers for all $n$ tasks with a union bound gives $T = O(nd/\rho m \log n)$, which decreases in the team size $\rho m$ and requires $\rho m > d$. *Remark:* the condition is deterministic and is validated directly in Appendix C. We are not aware of a comparable recovery condition for a persistent, per-observer mask, the regime where uniform-sampling matrix-completion guarantees do not apply; the finite-time rate under a strongly-exploiting (adaptive) policy is left open.

**Theorem 3 (collective speedup).** *(a)* At $\rho = 0$ robot $i$ sees only its own engaged entries $\langle p_i, u_j \rangle$ with $U$ unknown; for $d > 1$ these place no constraint on $\langle p_i, u_{j'} \rangle$ at an unseen $j'$, so the estimate is the prior (Proposition 1) and unseen skill is 0. *(b)* For $\rho > 0$ this is Theorem 2 specialised: its per-task coverage argument gives all-task recovery after $T_{\text{rec}} = O((nd/\rho m)\log n)$ rounds, decreasing as $1/m$; the only suppressed factor is the $\log n$ union bound over tasks. *Remark:* a corollary of Theorem 2; it is the formal counterpart of the value-of-broadcast and positive-scaling results in Section 6.3. The fold-in perturbation bound used above (cold-start error = basis-recovery + own-probe noise + ridge bias, exact at $k \geq d$) is stated in Appendix B.

## Appendix B. The fold-in error bound

For a newcomer factor $x_\star$ probed against an estimated basis $\hat{B} = B + \Delta$ ( $\|\Delta\| \leq \varepsilon$ ) with $k \geq d$ observations of noise $\sigma$ and ridge $\lambda$, the ridge fold-in prediction error splits, by a standard ridge-perturbation argument, cleanly into three sources, $\mathbb{E}|\hat{r} - r| \leq C_1 \varepsilon \|x_\star\|(1 + \|b\|/s) + C_2 \|b\| \sigma \sqrt{d}/s + C_3 \lambda \|x_\star\| \|b\|/s^2$ with $s = \sigma_{\min}(B)$, and is exact ($\hat{r} = r$) when $\varepsilon = \sigma = 0, \lambda \to 0, k \geq d$. It quantitatively explains the graceful degradation of cold-start skill as sensing becomes sparser.

## Appendix C. Empirical validation of the recovery condition (Theorem 2)

On the swarm's actual coverage patterns ( $m = 30, n = 240, d = 5, \rho = 0.5$ , noiseless to isolate identifiability), reconstructing each unseen pair $(i, j)$ from the observed entries by least squares gives error 0.000 exactly when robot $i$'s factor lies in the span of its visible engagers of $j$ (the condition of Theorem 2), versus a prior-floor reconstruction error of $\approx 0.30$ (an error, not a skill) otherwise, with graceful partial recovery as the spanning rank rises to $d$. The identifiability threshold is therefore exactly the spanning condition of Theorem 2.

## Appendix D. Reproducibility

All experiments use a block-model world with the signed inner-product reward of Section 3 and bootstrap 95% confidence intervals; the broadcast-rate, anytime, bake-off (Table 3), contention (Section 6.5), and scaling sweeps all use 16 random seeds; each reported number is averaged over the per-seed runs. The code and per-seed data needed to regenerate every figure and table are openly available (see Data availability).

**Hyperparameters.** Headline configuration: $m = 30$ robots, $n = 240$ tasks, true rank $d = 5$, guessed rank $\hat{d}$ drawn at random per run in $[d, 2d]$, horizon $T = 50$, offer size $c = 20$ (the unrestricted all-tasks menu, $c = n$, is the offer-size variant of Appendix F), own-observation noise $\sigma_{\text{own}} = 0.1$, broadcast-observation noise $\sigma_{\text{obs}} = 0.3$, persistent mask rate $\rho$ swept. SwarmCF: $\varepsilon$-greedy with $\varepsilon_0 = 0.5$ decaying by 0.93 per round to $\varepsilon_{\min} = 0.05$; ridge $\lambda = 10^{-2}$; 8 alternating-least-squares sweeps per refit; refit every 3 rounds; uniform observation weights (the noise level is not used). Identical exploration schedules and the same guessed rank $\hat{d}$ are given to every structured baseline for fairness; the choices are conservative for our claims (generous to baselines).

## Appendix E. The LatentSwarm environment

The LatentSwarm **package** (a modular, pip-installable Python package) is our simulator for ZK-MRTA (Section 6.5 adds capacity-1 contention). Its components are registered by name and selected by configuration: scenario builders (latent-trait generation), the environment, the policy library, the metrics, and the visualizations are separate pluggable modules, so a study is fully specified by one configuration object. It uses the Section 3 reward and observation model and adds capacity-1 contention, and SwarmCF enters only as one drop-in policy among the package's own (random, Hungarian oracle, MF-SGD, Independent-UCB).

**Traits and reward.** Each robot $i$ and task $j$ is assigned a hidden $d$-dimensional trait vector $(p_i, u_j)$ drawn from a shared signed Gaussian mixture over $K$ latent types (a block model), so the robot×task reward is low-rank of rank $d$, exactly as in Section 3. When robot $i$ engages task $j$ it earns the signed inner product $r_{ij} = \langle p_i, u_j \rangle$ (the Section 3 reward). The traits, not any spatial position, drive the reward; a t-SNE projection of the traits is used only for visualization.

**Decentralized, partial, private observation.** There is no communication. Each robot reads a public stream of engagement outcomes through a **persistent** per-pair mask $M_{ik} \sim \text{Bernoulli}(\rho)$ fixed for the mission ($M_{ii} = 1$): robot $i$ sees teammate $k$'s engagement only if $M_{ik} = 1$, and reads its outcome $\langle p_k, u_{a_k} \rangle + \eta_{ik}$ with **independent per-observer noise** $\eta_{ik} \sim \mathcal{N}(0, \sigma^2)$, so no two robots see the same stream. An optional action-identity channel may additionally corrupt which task a teammate is seen to select; it is disabled for the reported runs.

**Protocol and contention.** Each round every robot is offered a menu of tasks (all tasks by default; a random size-$c$ subset is an option) and selects one; engagements resolve in a random order under **capacity-1 contention** (only the first robot to pick a task that round succeeds; a colliding robot earns nothing). The regime is task-scarce ($n \gg T$) and a mission runs for $T$ rounds with no resets and no depletion. The guessed rank $\hat{d}$ is drawn at random per run and shared by every structured method, which also share the exploration schedule; SwarmCF is the weighted-ALS policy, maintaining its online low-rank factors from the per-observer broadcast and selecting the offered task it predicts best.

**Metrics. Earned (anytime) skill** normalizes the mean per-round reward to (policy − random)/(oracle − random), where the oracle is the per-round optimal one-to-one (Hungarian) matching of robots to distinct tasks, the centralized capacity-1 ceiling. **Unseen-pair skill** measures each robot's decision quality on tasks it never engaged, from its learned model, self-normalized so the oracle is 1 and random is 0. The configuration is $m = 30$, $n = 240$, $d = 5$, $\hat{d} \sim \text{Uniform}\{d, \ldots, 2d\}$, all $n$ tasks offered, $T = 50$, $\rho = 0.5$, $\sigma = 0.3$, with 16 seeds and bootstrap 95% confidence intervals.

Algorithm 3: LatentSwarm mission (a ZK-MRTA instantiation with capacity-1 contention; each robot runs a policy, e.g. SwarmCF)

draw signed low-rank traits $p_i$ (robots), $u_j$ (tasks) from a shared $K$-type Gaussian mixture (rank $d$)
draw a persistent broadcast mask $M_{ik} \sim \text{Bernoulli}(\rho)$, $M_{ii} = 1$ (fixed for the mission)
draw a guessed rank $\hat{d} \sim \text{Uniform}\{d, \ldots, 2d\}$, shared by all structured methods
for round $t = 1, \ldots, T$:
  offer each robot $i$ a menu $S_{it}$ (all tasks by default, or a random size-$c$ subset)
  each robot $i$ selects $a_i \in S_{it}$ from its policy
  taken ← ∅; for each engaging robot $i$, in random order:
    if $a_i \in$ taken (capacity-1): robot $i$ earns 0 (collision); continue

robot $i$ earns $r = \langle p_i, u_{a_i} \rangle$; taken ← taken ∪ $\{a_i\}$

broadcast: each robot $i$ records, for visible teammates $k$ ($M_{ik} = 1$),

$(k,\ a_k, \langle p_k, u_{a_k} \rangle + \eta_{ik})$, private $\eta_{ik} \sim \mathcal{N}(0, \sigma^2)$

score: earned skill vs random / Hungarian oracle; unseen-pair skill on never-engaged tasks

## Appendix F. Robustness to the offer size, the masking model, and approximate low-rank

Two evaluation choices in the body are robustness knobs rather than load-bearing assumptions: the offer size and the masking model. We vary each here and find the categorical separation and the masking-robustness of SwarmCF unchanged.

**Offer size.** The body restricts each offer to a uniform random size-$c$ subset ($c = 20$); here we instead offer every robot all $n$ tasks each round (the unrestricted menu, $c = n$). Figure 9 repeats the masking sweep under both menus. The categorical result is unchanged: every low-rank method stays well above the structure-free floor at both menu sizes (Figure 9a), so the structure-versus-no-structure separation, the paper's main claim, does not depend on the offer size. What changes is the ordering *among* low-rank methods. Under the size-$c$ menu the per-round random offers force each robot to engage a varied set of tasks, giving the online estimator the coverage it needs, so SwarmCF leads the batch completers; under the all-tasks menu every robot instead selects greedily over all $n$ tasks, the collective engagement coverage narrows, and SwarmCF's unseen-pair skill falls to or below the batch and probe-based methods, whose scheduled exploration is insensitive to the menu. The advantage of an online greedy estimator over batch completion is therefore contingent on the offered menu supplying exploration diversity; a probe phase (the hybrid variant) or a non-greedy menu restores it. This is a limitation of the greedy online policy, not of the low-rank approach, whose categorical advantage over structure-free learning is unaffected. More generally, replacing the uninformed $\varepsilon$-greedy probe with confidence-directed exploration, a Bayesian posterior over the factors that probes where the shared structure is least pinned down, restores coverage without a fixed probe budget; we develop this in a follow-up.

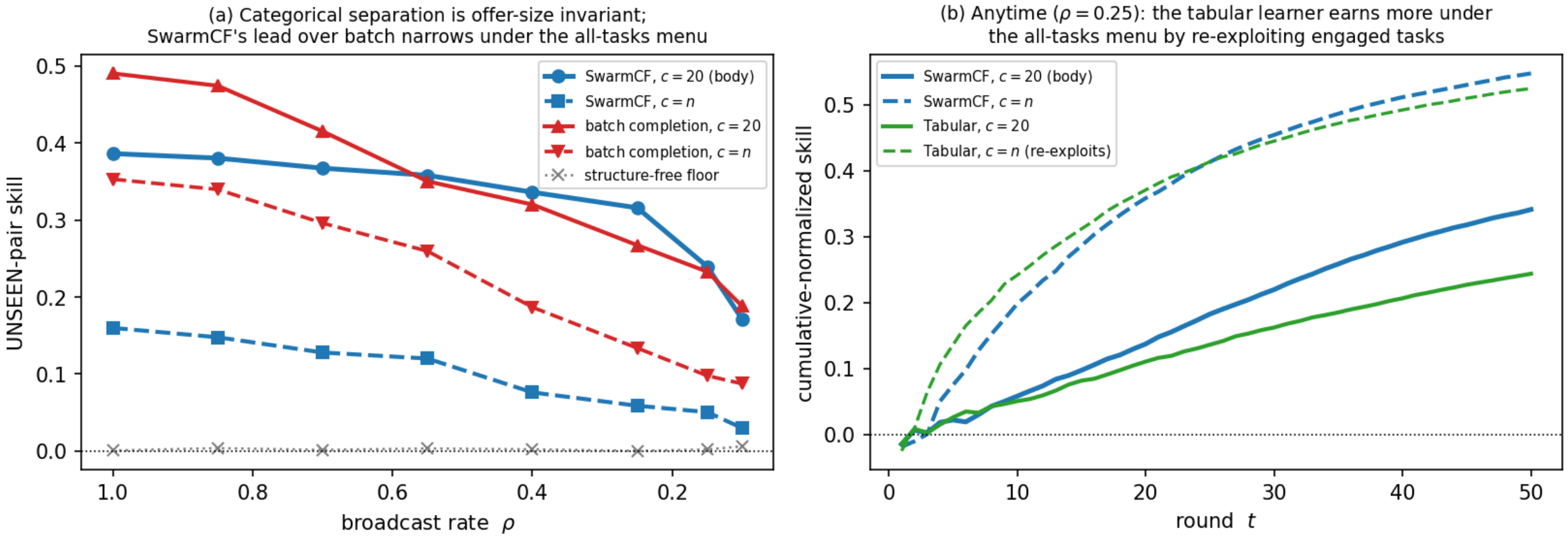

*Figure 9. Offer-size robustness. (a) Unseen-pair skill versus broadcast rate $\rho$ under the body's size-$c$ menu ($c = 20$) and the unrestricted all-tasks menu ($c = n$), for SwarmCF and the batch completer (the structure-free floor is shown once): every low-rank method stays above the floor at both menu sizes (the categorical separation is offer-size invariant), but SwarmCF's lead over the batch method narrows under the all-tasks menu, where greedy selection reduces engagement diversity. (b) Anytime cumulative-reward skill ($\rho = 0.25$): the tabular learner earns more under the all-tasks menu because it can re-exploit any task it has already engaged, whereas the scarce menu makes re-exploitation rare.*

**Masking model.** The body uses a *persistent* observation mask (each pair $(i, k)$ is visible or not for the whole mission). An *i.i.d.* per-round mask redraws visibility every round, which reduces to standard uniform sub-sampling of the broadcast. Figure 10 compares the two. Final-policy unseen skill and anytime skill are essentially invariant to the masking model, as Theorem 3 predicts; what differs is decentralization durability: under the persistent mask each robot keeps a permanently different view, so the robots' learned models stay distinct (state-uniqueness is durable), whereas under the i.i.d. mask the blind spots average out over rounds and the models converge (state-uniqueness is transient, decreasing with the horizon). The persistent mask is therefore the harder, more realistic regime, and is the one used throughout the body.

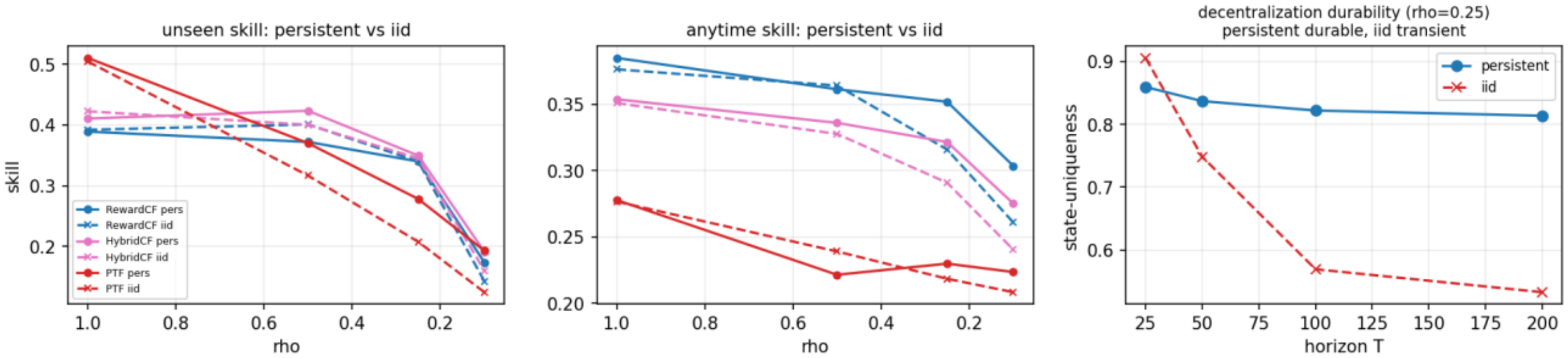

*Figure 10. Persistent versus i.i.d. masking. Unseen-pair and anytime skill are essentially invariant to the masking model (left, center); decentralization (state-uniqueness of the robots' learned models) is durable under the persistent mask but transient under the i.i.d. mask, which averages out blind spots over the mission (right).*

**Theorem 1's strict regime.** The body's anytime comparison (Figure 3) uses a size-$c$ menu with $cT/n \approx 4$, outside the strict scarce-offer regime $cT = o(n)$ in which Theorem 1 predicts the structure-free anytime collapse. Figure 11a re-runs the anytime comparison in that strict regime ($c = 3$, $cT/n < 1$): the structure-free learners' cumulative skill collapses to the floor exactly as Theorem 1 states, while SwarmCF still earns, so the theorem operates as predicted when its hypothesis holds.

**Restoring the online lead under the all-tasks menu.** The narrowing of SwarmCF's lead over batch completion under the all-tasks menu (Figure 9) is a property of uninformed greedy exploration, not of the online estimator: adding a short UCB probe phase (a hybrid that probes, then runs online ALS) restores the lead at $c = n$ across broadcast rates (Figure 11b). The contingency is the exploration schedule, which a probe, or the confidence-directed exploration of the follow-up, repairs.

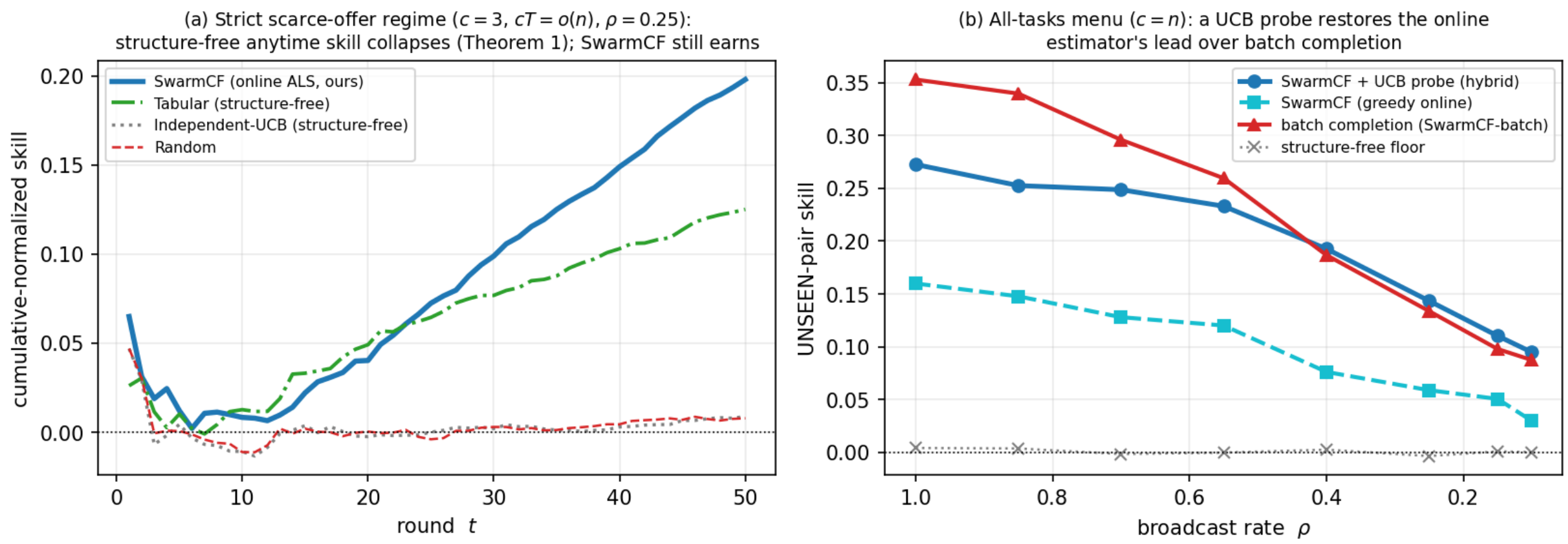

*Figure 11. Two Appendix-F ablations. (a) Strict scarce-offer regime ($c = 3$, $cT = o(n)$, $\rho = 0.25$): structure-free learners' cumulative skill collapses to the floor (Theorem 1) while SwarmCF still earns. (b) Under the all-tasks menu ($c = n$), a short UCB probe restores the online estimator's unseen-pair lead over batch completion that pure greedy selection gives up (structure-free floor shown for reference). Means over 16 seeds.*

**Approximate low-rank.** The categorical separation assumes the reward is low-rank; we test how it degrades when the reward is only *approximately* so. We perturb the rank-$d$ block reward with a full-rank Gaussian term, $R_\varepsilon = (R + \varepsilon\, s\, G)/\sqrt{1+\varepsilon^2}$ ($G_{ij} \sim \mathcal{N}(0,1)$, $s = \mathrm{std}(R)/\mathrm{std}(G)$), which holds the entry-wise scale fixed (so the observation SNR is unchanged) while moving energy out of the rank-$d$ subspace: the low-rank energy fraction is $1/(1+\varepsilon^2)$ and the effective rank rises from $d$ toward $\min(m,n)$ as $\varepsilon$ grows. Figure 12 sweeps $\varepsilon$ at the masked headline $\rho = 0.25$. SwarmCF degrades **gracefully**: unseen-pair skill falls from 0.32 at $\varepsilon = 0$ (effective rank 5) to 0.20 at $\varepsilon = 0.5$ (effective rank about 28; roughly 80% of the reward energy still low-rank) and 0.08 at $\varepsilon = 1$ (about 50%), staying above the structure-free floor (about 0 at every $\varepsilon$, intervals straddling zero). The advantage is therefore a property of *exploitable* low-rank structure, not of exact low-rankness: it weakens smoothly as structure is destroyed and reaches the floor only when essentially none remains.

***Figure 12.*** *Approximate-low-rank robustness: unseen-pair skill versus a full-rank perturbation of strength $\varepsilon$ (top axis: the resulting effective rank at 99% energy), at the masked headline $\rho = 0.25$. SwarmCF and its batch variant degrade gracefully as the reward leaves the rank-d subspace and approach the structure-free floor only when little low-rank structure remains; the structure-free learner is at the floor for every $\varepsilon$. Means with bootstrap 95% CIs over 16 seeds.*